\pdfoutput=1
\documentclass[11pt]{article}
\PassOptionsToPackage{table}{xcolor}

\usepackage[final]{acl}
\usepackage{times}
\usepackage{latexsym}
\usepackage[T1]{fontenc}
\usepackage[utf8]{inputenc}
\usepackage{microtype}
\usepackage{etoolbox}
\usepackage{ragged2e}

\makeatletter
\apptocmd{\thebibliography}{%
  \RaggedRight
  \hyphenpenalty=50
  \exhyphenpenalty=50
}{}{}
\makeatother

\usepackage{inconsolata}
\usepackage{graphicx}
\usepackage{booktabs}
\usepackage{multirow}
\usepackage{pifont}
\newcommand{\cmark}{\ding{51}}
\newcommand{\xmark}{\ding{55}}

\usepackage{listings}
\usepackage{xcolor}
\usepackage[most]{tcolorbox}
\tcbuselibrary{listings,skins,breakable,fitting}

\usepackage{colortbl}
\usepackage{arydshln}

\newcommand{\tri}{\ensuremath{\triangle}}
\newcommand{\TRI}{\cellcolor{green!15}\tri}
\newcommand{\X}{\cellcolor{blue!10}\xmark}

\lstdefinelanguage{ebnf}{
  morecomment=[s]{(*}{*)},
  morestring=[b]",
  morestring=[b]',
}
\usepackage{subcaption}
\usepackage{float}
\usepackage{needspace}
\usepackage{stfloats}
\newcommand{\std}[1]{$_{\pm#1}$}

\definecolor{ltempurple}{HTML}{6D28D9}
\definecolor{ltemblue}{HTML}{4995B5} 
\colorlet{ltemmerge}{ltemblue!18}
\newcommand{\ltemmk}[1]{\textcolor{ltemblue!80!black}{\textbf{#1}}}
\lstdefinestyle{ltemsrc}{basicstyle=\ttfamily\footnotesize, columns=fixed, basewidth=4.5pt, frame=none,
  backgroundcolor=\color{gray!8}, escapeinside={(*}{*)}, aboveskip=0pt, belowskip=0pt, breaklines=false}
\title{\raisebox{-0.17em}{\includegraphics[height=1.05em]{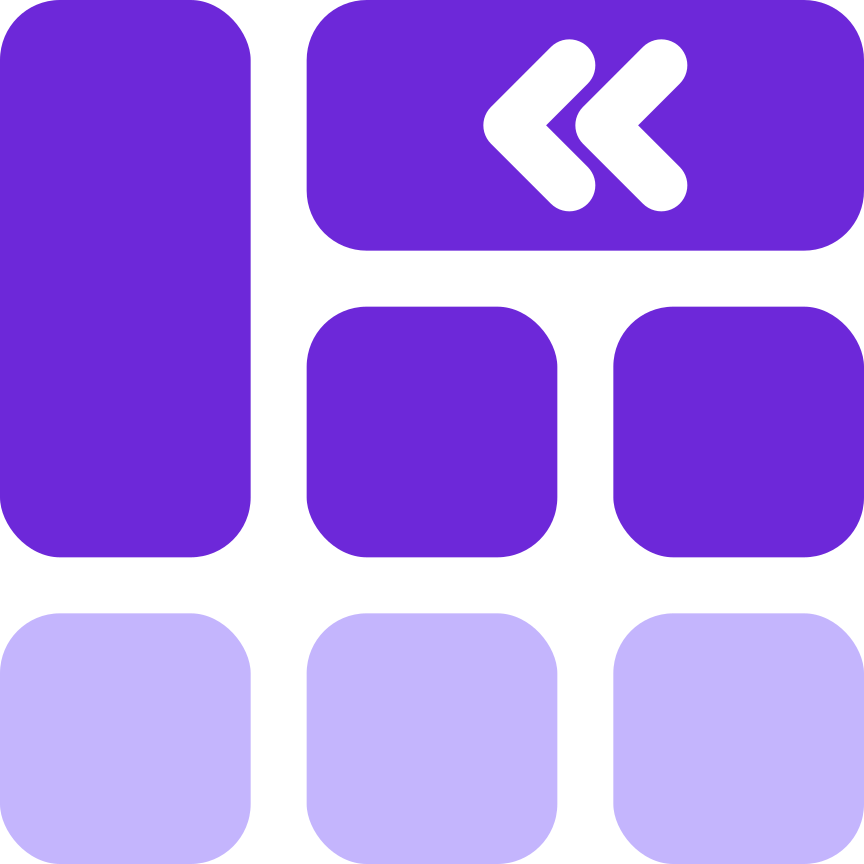}}\hspace{0.35em}\textcolor{ltempurple}{LTEM}: LLM-Friendly Table-Enhanced Markdown for Table Understanding and Generation}

\providecommand{\LTEMAuthorSeparator}{,\ }
\author{Eunbin Kim$^{1}$\LTEMAuthorSeparator Hyeonseok Kang$^{1}$\LTEMAuthorSeparator Sangkeun Jung$^{1,2}$\thanks{Corresponding author.} \\
  $^{1}$Department of Computer Science and Engineering, \\
  Chungnam National University, Republic of Korea \quad $^{2}$EurekaAI\thanks{EurekaAI: \url{https://eurekax.co/}} \\
  \texttt{\{eunbinkim777,dnfldjaak11,hugmanskj\}@gmail.com}}

\begin{document}
\maketitle
\begin{abstract}
Tables in real-world documents, including scientific papers, financial reports, and statistical summaries, frequently use merged cells, multi-level headers, and sectioned layouts. Yet the lightweight formats LLMs handle best (Markdown, CSV) cannot express these structures, while expressive formats (HTML, LaTeX) impose excessive token cost and parsing complexity. To enable systematic analysis of this trade-off, we introduce a \textbf{taxonomy} decomposing table structure into six atomic features (F1--F6) and corresponding complexity levels (L1--L6), and construct \textbf{STB} (Synthetic Table Benchmark), a controlled benchmark with 1,089 tables systematically distributed across complexity levels. Using this framework, we propose \textbf{LTEM} (LLM-friendly Table-Enhanced Markdown), a formally specified, minimal extension of Markdown that explicitly encodes spans and hierarchical headers. Experiments across five state-of-the-art LLMs show that LTEM achieves accuracy comparable to HTML, offers a favorable accuracy--token trade-off relative to HTML and LaTeX, and produces substantially fewer invalid responses than LaTeX; for table generation, LTEM is comparable to MultiMarkdown though HTML is best. We release our taxonomy, benchmark, and format specification to support future research on LLM-table interaction.
\end{abstract}

\section{Introduction}

LLMs excel at natural language tasks, yet many practical applications, including document editing, report writing, and data summarization, require tables. Because tables are two-dimensional while LLM inputs are linear, models must reconstruct header roles, spans, and hierarchies from serialized text, making performance sensitive to the \textit{representation format}.

Recent work confirms that format choice matters: identical tables yield different performance depending on serialization, delimiters, and ordering \citep{sui2024tablemeetsllm,li2025longtablebench,wu2025realhitbench}. However, existing studies provide limited tools for comparing \textit{what structural information} each format preserves.

Real-world tables often include merged cells, multi-level headers, and sectioned layouts. Lightweight formats (Markdown, CSV) are easy for LLMs but cannot faithfully express these structures, while expressive formats (HTML, LaTeX) are verbose and increase token cost and parsing burden. For example, GitHub-Flavored Markdown (GFM) defines pipe-delimited tables \citep{githubdocs_markdown_tables} but does not support \texttt{colspan}/\texttt{rowspan} \citep{mdn_gfm_markdown}. MultiMarkdown adds column-spanning syntax through repeated pipes \citep{multimarkdown_tables}. In contrast, HTML and LaTeX can represent complex tables but require substantial markup interpretation.

To systematize this trade-off, we define six \textbf{Structural Features (F1--F6)} and \textbf{Complexity Levels (L1--L6)} as cumulative feature requirements. This taxonomy makes table complexity explicit and enables direct comparison of formats by expressiveness.

We then propose \textbf{LTEM}\footnote{Project page: \url{https://hugman.github.io/ltem/}} (LLM-friendly Table-Enhanced Markdown), a minimal extension of Markdown that explicitly encodes spans and multi-level headers while remaining linear to read and parsable in a single pass. LTEM targets L3--L5 structures with less verbosity than HTML/LaTeX, aiming to improve token efficiency and response stability without sacrificing expressiveness.

We evaluate LTEM on \textbf{Table Understanding} (Table QA) and \textbf{Table Generation} across formats (Markdown, CSV, HTML, LaTeX, MultiMarkdown, LTEM) and five state-of-the-art LLMs, and analyze \textbf{Token Efficiency} using token-normalized metrics.

Our contributions are as follows:
\begin{itemize}
  \item \textbf{Structural complexity taxonomy}: We formalize table difficulty and format expressiveness via F1--F6 and L1--L6, providing a systematic framework for comparing serialization capabilities.
  \item \textbf{STB benchmark}: We construct a controlled benchmark with 1,089 tables systematically distributed across complexity levels, enabling controlled analysis of structural effects on LLM performance.
  \item \textbf{LTEM}: We propose a minimal, formally specified Markdown extension that instantiates the taxonomy-derived requirements, explicitly encoding cell spans and hierarchical headers while preserving LLM-friendly linear readability.
  \item \textbf{Empirical trade-off analysis}: We evaluate understanding, generation, and token efficiency across formats and diverse LLM backbones.
\end{itemize}

\section{Related Work}

\subsection{Table QA and Table Reasoning}
Early Table QA focused on semantic parsing, mapping questions to executable programs. WTQ \citep{pasupat-liang-2015-compositional} and WikiSQL \citep{zhong2017seq2sql} established QA over Wikipedia tables, and later work introduced pretrained table encoders such as TAPAS and TaBERT \citep{herzig-etal-2020-tapas,yin-etal-2020-tabert}. However, these datasets largely emphasize flat tables, underrepresenting document-style structures such as merged cells and hierarchical headers. Benchmarks including HiTab and TAT-QA better reflect hierarchical and numerically grounded reasoning \citep{cheng-etal-2022-hitab,zhu-etal-2021-tat}. In contrast to work centered on tasks or model variants, we treat \textbf{serialization format} as a primary variable and study its impact on stability and token efficiency across structural complexity.

\subsection{LLMs on Tables and Format Choice}
Recent studies show that LLM performance on tabular tasks is sensitive to serialization, delimiters, and ordering \citep{sui2024tablemeetsllm,fang2024large,lu2024tableprocesssurvey}. LongTableBench emphasizes format and context length in long-context table reasoning \citep{li2025longtablebench}, and RealHiTBench compares representations for complex hierarchical tables across LaTeX, HTML, and images \citep{wu2025realhitbench}. While these works establish that format matters, they offer limited tools to compare \emph{what structural information} each representation preserves. Our taxonomy addresses this gap by decomposing table structure into atomic operations.

\subsection{Markup Expressiveness and Text-to-Table Evaluation}
Markdown tables are common: GFM defines pipe-delimited tables \citep{githubdocs_markdown_tables} but lacks structural attributes such as \texttt{colspan}/\texttt{rowspan} \citep{mdn_gfm_markdown}, while MultiMarkdown adds column-spanning syntax through repeated pipes \citep{multimarkdown_tables}. HTML and LaTeX provide explicit span constructs for complex tables \citep{whatwg_html_tables,latexref_multicolumn,ctan_multirow}. Their more verbose syntax motivates our evaluation of format-dependent accuracy and structural consistency. For text-to-table generation, evaluation is difficult because multiple surface forms can represent the same structure. Prior work proposes multi-granularity table metrics \citep{pal2023multitabqa}, statement-based checks \citep{ramu2024tabeval}, and discusses structure-aware metrics under constrained decoding \citep{oestreich2025evaluating}. We analyze format-dependent structural consistency via normalized representations, following exact-match perspectives common in structured output evaluation \citep{yu2018spider}.

\section{Taxonomy of Table Structural Complexity}

An objective comparison of table representation format capabilities requires clear criteria. This section presents a \textbf{taxonomy} that serves as an evaluation methodology for systematically characterizing table structural complexity and the structural information a serialization must preserve for LLMs. This classification scheme comprises two axes: \textbf{Structural Features (F1--F6)}, the atomic operations for expressing table structure, and \textbf{Complexity Levels (L1--L6)}, the cumulative feature requirements for expressing specific tables.

\subsection{Structural Features}

We define six atomic operations for expressing table structure. Each Feature is independent of the others, enabling binary assessment of whether a specific structural capability is present or absent.

\paragraph{F1 (Grid Formation).}
F1 (Grid Formation) represents the most basic requirement: that cells ``exist'' at row-column intersections.

\paragraph{F2 (Header Designation).}
F2 (Header Designation) indicates that cells have assigned ``roles.'' In HTML, header cells use \texttt{<th>} while data cells use \texttt{<td>}; in Markdown, rows above the separator line (\texttt{|-{}-{}-|}) are treated as headers.

\paragraph{F3 (Cell Merging).}
F3 (Cell Merging) is the critical branching point in this taxonomy. A substantial portion of tables in scientific papers require cell merging for group headers or category representation.

\paragraph{F4 (Header Nesting).}
F4 (Header Nesting) applies when headers themselves have parent--child hierarchies, exemplified by grouping quarterly revenue as ``Q1'', ``Q2'', ``Q3'' under a ``Revenue'' superheader.

\paragraph{F5 (Section Stacking).}
F5 (Section Stacking) and F6 (Table Composition) address compound table structures. F5 applies when multiple sections sharing identical column structure (schema) coexist within a single table, such as displaying ``Model A Results'' and ``Model B Results'' in one table in regression analysis.

\paragraph{F6 (Table Composition).}
F6 applies when tables with different schemas constitute a single logical unit, such as placing ``(a) Correlation Matrix'' and ``(b) Sample Sizes'' in one figure.

\begin{table*}[t]
\centering
\small
\setlength{\tabcolsep}{4.5pt}
\renewcommand{\arraystretch}{1.0}

\begin{tabular}{lccccc:cc}
\toprule
\textbf{Feature} & \textbf{CSV} & \textbf{MD} & \textbf{MMD} & \textbf{HTML} & \textbf{LaTeX} & \textbf{LTEM} & \textbf{LTEM Syntax} \\
\midrule
F1 Grid Formation     & \cmark & \cmark & \cmark & \cmark & \cmark & \cmark & \texttt{| |} \\
F2 Header Designation & \TRI   & \cmark & \cmark & \cmark & \cmark & \cmark & \texttt{|-{}-{}-|} \\
F3 Cell Merging       & \X     & \X     & \TRI   & \cmark & \cmark & \cmark & \texttt{<<}, \texttt{\textasciicircum\textasciicircum}, \texttt{\{c:N,r:M\}} \\
F4 Header Nesting     & \X     & \X     & \TRI   & \cmark & \cmark & \cmark & multi-header rows \\
F5 Section Stacking   & \X     & \X     & \X     & \TRI   & \cmark & \cmark & section-label rows \\
F6 Table Composition  & \X     & \X     & \X     & \TRI   & \cmark & \cmark & \texttt{| \# id |}, \texttt{@ref} \\
\midrule
\end{tabular}
\caption{Structural feature support by format. \cmark~= native support; $\triangle$~= partial or non-standard; \xmark~= not supported. Dashed vertical rule separates existing formats from LTEM.}
\label{tab:ltem_feature_support}
\end{table*}

\subsection{Complexity Levels}

Complexity Level is defined as the \textbf{cumulative features required} to express a specific table. Each Level includes all Features from previous Levels and adds new required Features. Complexity Level definitions are summarized in Table~\ref{tab:complexity_levels} in the Appendix.

The key implication of this cumulative structure is: a format that cannot support a specific Level cannot structurally express any higher Levels. For example, a format not supporting F3 (Cell Merging) cannot structurally express not only L3 but also L4, L5, and L6 tables.

\subsection{Expressiveness of Existing Formats}

We apply the above taxonomy to analyze the expressiveness of major table representation formats. The analysis covers five formats spanning minimal data exchange formats (CSV), lightweight markup with high frequency in LLM training data (Markdown, MultiMarkdown), and document standards with complete structural expressiveness (HTML, LaTeX). Two important patterns emerge from this analysis (Table~\ref{tab:ltem_feature_support}).

\textbf{First}, F3 (Cell Merging) is the decisive branching point. Markdown does not support F3 at all, fundamentally precluding expression of L3+ tables. MultiMarkdown supports \texttt{colspan} in some renderers but lacks standardization, with even more limited \texttt{rowspan} support. This is not merely a feature absence but a structural limitation: the syntax for encoding this information simply does not exist.

\textbf{Second}, only LaTeX has complete expressiveness. HTML can approximate F5 and F6 using multiple \texttt{<tbody>} elements or separate \texttt{<table>} elements, but cannot structurally specify logical relationships between sections. LaTeX is the only format that can express structures at all levels through a nesting-based approach using nested \texttt{tabular} environments and \texttt{subtable}.

This analysis presents an important implication overlooked in prior research: the format LLMs process most efficiently (Markdown) cannot express L3+ tables commonly found in scientific papers, while formats capable of such expression (LaTeX, HTML) are inefficient for LLM processing. 

LaTeX's \texttt{\textbackslash multicolumn\{2\}\{c\}\{...\}} and HTML's \texttt{<td colspan="2">} encode cell merging information, but LLMs must undergo additional steps to interpret this syntax and infer actual cell positions and ranges. The same burden exists when providing tables as images, requiring inference of cell boundaries and merging relationships from visual structure. 

This taxonomy-based analysis yields concrete serialization requirements for LLM table processing: a format targeting L3--L5 structures should explicitly encode span information and header hierarchies while maintaining linear readability. This motivates the design of a new table markup language optimized for LLM comprehension.

\section{LTEM: Language Design}
To demonstrate how the taxonomy-derived requirements (Section~3) translate into concrete format design, we present LTEM (\textbf{L}LM-friendly \textbf{T}able-\textbf{E}nhanced \textbf{M}arkdown) as an instantiation that explicitly encodes F3--F5 features while preserving Markdown's linear readability. This section describes design principles, syntax specification, and LTEM's position within the landscape presented in Section~3.

\subsection{Design Principles}

LTEM's design is grounded in \textit{five} principles derived from the taxonomy-based analysis in Section~3, each addressing specific structural requirements identified by our framework.

\paragraph{Markdown Compatibility.}
LLM training corpora include substantial proportions of sources using Markdown format, such as GitHub and StackExchange (e.g., GitHub 4.5\%, StackExchange 2\% in LLaMA; \citep{touvron2023llama}). This composition leads LLMs to exhibit high familiarity with Markdown syntax; in fact, major commercial LLMs such as ChatGPT and Claude use Markdown as their default output format for table generation. LTEM is designed as a superset of Markdown to leverage this existing knowledge. Even when new syntax elements render as plain text in existing Markdown parsers, basic table structure is preserved. This design enables LLMs to understand LTEM as an extension of familiar Markdown rather than learning an entirely new format.

\paragraph{Linear Readability.}
Table structure should be \textbf{semantically comprehensible} when reading source text sequentially from top to bottom. This principle aligns with LLMs' autoregressive processing and simultaneously facilitates human review and debugging.

\paragraph{Explicit Structure.}
Structural information should be expressed through explicit markers rather than implicit rules. For example, a cell with colspan=2 is clearly indicated with a \texttt{\{c:2\}} attribute or \texttt{<<} marker, eliminating the burden of contextual inference by LLMs. This blocks a major source of structural hallucination.

\paragraph{Minimal Grammar Extension.}
We minimize the number of new syntax elements. LTEM adds only seven syntax elements to standard Markdown: four for span expression (\texttt{<<}, \texttt{\textasciicircum\textasciicircum}, \texttt{\{c:N\}}, \texttt{\{r:N\}}), one for in-cell line breaks (\texttt{\textbackslash\textbackslash}), one for table identification (\texttt{| \# id |}), and one for references (\texttt{@}). Appendix~\ref{app:ltem_design_ablation} reports a design-choice ablation comparing alternative forms for five LTEM elements.

\paragraph{Unidirectional Parsing.}
While Linear Readability concerns semantic understanding, this principle concerns \textbf{syntax parsing implementation}. LTEM is parsable in a single pass without forward references, guaranteeing efficient parser implementation with linear time complexity.

\subsection{Core Syntax}
\label{sec:ltem-core-syntax}

LTEM's core syntax addresses cell spanning and multi-level headers. Appendix~\ref{app:rendering} shows the tables these constructs render to.

\begin{figure}[t]
  \centering

  \begin{subfigure}[t]{\columnwidth}
    \centering
    \includegraphics[width=\columnwidth,height=1.5cm,keepaspectratio]{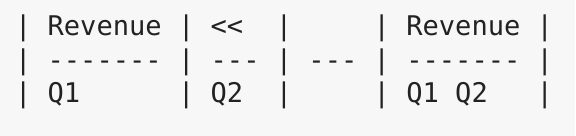}
    \caption{Motivation: span-aware vs.\ span-unaware.}
    \label{fig:ltem-colspan-a}
  \end{subfigure}

  \begin{subfigure}[t]{\columnwidth}
    \centering
    \includegraphics[width=\columnwidth,height=1.5cm,keepaspectratio]{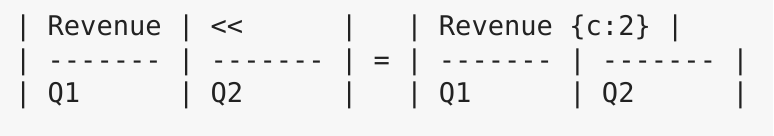}
    \caption{Equivalent encodings: \texttt{<<} vs.\ \texttt{\{c:N\}}.}
    \label{fig:ltem-colspan-b}
  \end{subfigure}

  \caption{Colspan in LTEM. (a) illustrates why explicit span markers matter; (b) shows marker- and attribute-based forms are structurally equivalent.}
  \label{fig:ltem-colspan}
\end{figure}

\paragraph{Colspan.}
Column spanning is expressed via marker syntax (\texttt{<<}) or attribute syntax (\texttt{\{c:N\}}). The two forms are structurally equivalent, as illustrated in Figure~\ref{fig:ltem-colspan}.

Marker syntax maintains visual alignment, while attribute syntax is efficient for large span values, allowing concise expression like \texttt{\{c:5\}} without repeated marker entry.

\paragraph{Rowspan.}
Row spanning is expressed via the \texttt{\textasciicircum\textasciicircum} marker or \texttt{\{r:N\}} attribute. Compound spanning combines as \texttt{\{c:N,r:M\}}.

\begin{figure}[t]
  \centering
  \includegraphics[width=\columnwidth,height=1.5cm,keepaspectratio]{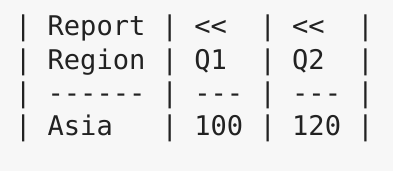}
  \caption{Multi-level headers in LTEM: multiple header rows above the separator line naturally encode hierarchical header nesting (F4).}
  \label{fig:ltem-multiheader}
\end{figure}

\paragraph{Multi-header Rows.}
All rows above the separator line (\texttt{|-{}-{}-|}) are treated as headers. This design naturally supports F4 (Header Nesting) without additional syntax (Figure~\ref{fig:ltem-multiheader}).

\paragraph{Multiline Cell.}
The \texttt{\textbackslash\textbackslash} escape sequence represents line breaks within cells.

\subsection{Rich Cell Content}

\begin{figure}[t]
  \centering
  \includegraphics[width=\columnwidth]{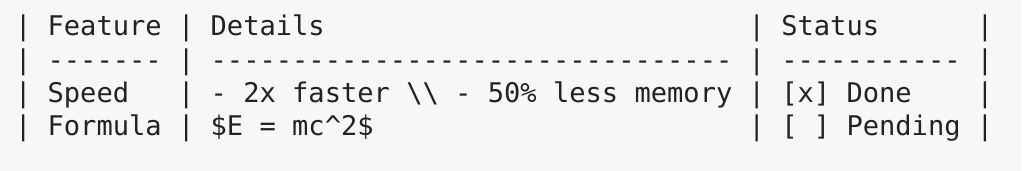}
  \caption{Rich cell content in LTEM: inline line breaks (\texttt{\textbackslash\textbackslash}) enable structured content (lists, checkboxes) and math within a single cell.}
  \label{fig:ltem-richcell}
\end{figure}

LTEM supports structured content within cells based on \texttt{\textbackslash\textbackslash}. Bullet lists, numbered lists, checkboxes, and formulas can be combined within a single cell, as illustrated in Figure~\ref{fig:ltem-richcell}.

\subsection{Relational Tables}
\begin{figure}[t]
  \centering
  \includegraphics[height=5cm,keepaspectratio]{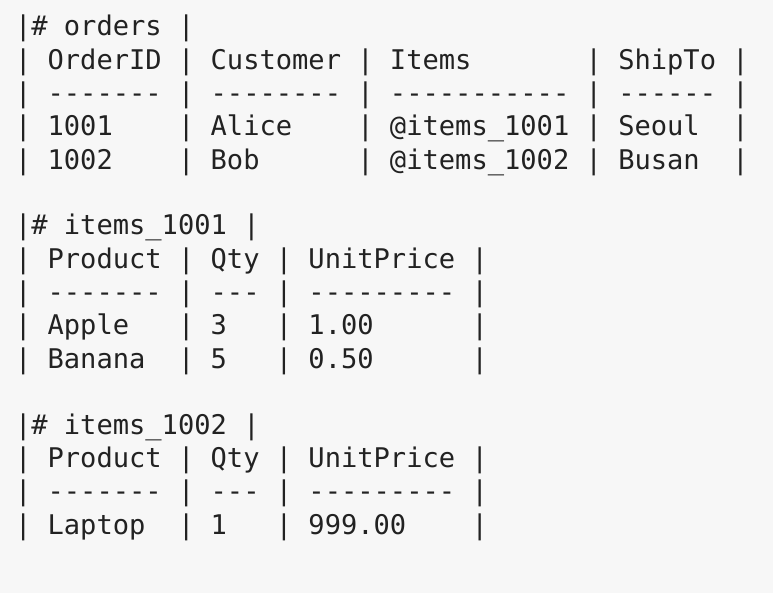}
  \caption{Relational tables in LTEM: tables are decomposed and linked via explicit table IDs (\texttt{| \# id |}) and references (\texttt{@ref}) instead of nesting.}
  \label{fig:ltem-relational}
\end{figure}

LaTeX and HTML permit nested structures inserting tables within cells, but this requires recursive parsing, disrupts linear reading, and expresses inter-table relationships only implicitly. To address this, LTEM adopts an alternative approach: \textbf{table decomposition with explicit references}, as illustrated in Figure~\ref{fig:ltem-relational}. This is analogous to normalization principles in relational databases.

A table is identified by a unique label placed in the first row as \texttt{| \# identifier |}. Other tables are referenced with the \texttt{@} prefix.

\subsection{Formal Specification}

LTEM syntax is defined as a Context-Free Grammar (CFG). The complete EBNF (Extended Backus-Naur Form) specification is provided in Appendix~\ref{app:ltem-ebnf}, together with syntax validation rules (V1--V12) and AST and TableStructure JSON schemas.

\section{Experimental Setup}

\subsection{Experimental Design}

The experiments are designed to answer the following research questions:

\begin{itemize}
    \item \textbf{RQ1}: Can LTEM sufficiently express diverse tables?
    \begin{itemize}
        \item \textbf{RQ1-1}: Does LTEM improve LLM table comprehension?
        \item \textbf{RQ1-2}: Does LTEM achieve high fidelity in table generation?
    \end{itemize}
    \item \textbf{RQ2}: Is LTEM more token-efficient than existing formats?
\end{itemize}

To address these questions, we utilize two complementary datasets. First, the Synthetic Table Benchmark (STB) provides synthetic tables with systematically controlled structural complexity, enabling controlled analysis of format effects at each Complexity Level. Second, existing public benchmarks such as HiTab~\cite{cheng-etal-2022-hitab} ensure comparability with prior research.

\subsection{Synthetic Table Benchmark (STB)}

Existing table benchmarks were collected for specific tasks like Table QA or Table-to-Text, without systematic control of table structural complexity. Most tables in WikiTableQuestions have simple header--data structures (L2), with only a minority containing cell merging or multi-level headers. This distributional bias implies that conclusions like ``format A outperforms format B'' may only hold within specific complexity ranges.

To overcome this limitation, we constructed the Synthetic Table Benchmark (STB), a synthetic dataset systematically distributed according to the Complexity Levels (L1--L6) defined in Section~3. STB comprises 1{,}089 tables and 4{,}356 question--answer pairs, with each table containing an average of 4 QA pairs.

STB's complexity distribution was deliberately designed for research purposes (Table~\ref{tab:stb-level}). Notably, approximately 30\% is allocated to L3, the first level requiring Cell Merging and the boundary where Markdown and CSV become structurally incapable of representation. We anticipated LTEM's differential value would be most evident at this point. To ensure content realism, we defined five domains: \textit{Scientific, Financial, Product, Survey}, and \textit{Schedule} (Table~\ref{tab:stb-domain}), with complexity levels and domains sampled independently to prevent specific domains from being biased toward specific complexity levels.

Table generation follows a 3-stage pipeline: (i) rule-based determination of table structure (row/column counts and merging patterns), (ii) Claude Sonnet~4.5 and Opus~4.5~\citep{anthropic2025sonnet45,anthropic2025opus45} populate domain-appropriate headers and cell values, then (iii) GPT-5-mini generates question--answer pairs~\citep{openai2025gpt5systemcard}. Question types comprise LOOKUP, COUNTING, COMPARISON, and AGGREGATION, referencing operation types repeatedly used in prior Table QA research~\cite{pasupat-liang-2015-compositional,cheng-etal-2022-hitab}. Additional implementation details and prompt templates are provided in Appendices~\ref{app:stb_pipeline} and~\ref{app:prompts}.

Since automatically generated QA may contain errors, we adopted an indirect verification approach using \textbf{model agreement} rather than explicitly computing or verifying answer labels with rules. Specifically, we input identical tables and questions to two models from different families (Qwen3-VL-8B-Instruct and Phi-4-mini-instruct~\citep{bai2025qwen3vl,abdin2025phi4mini}), retaining QA samples only when both models' outputs match the GPT-5-mini-generated answer under the same normalization criteria. This agreement serves as a reliability filter, although errors shared across models may remain.

\subsection{Table QA Benchmarks}
To verify whether trends observed in the synthetic dataset (STB) generalize to real-world tables, we used HiTab~\cite{cheng-etal-2022-hitab}, a public benchmark for hierarchical table question answering. HiTab contains tables extracted from Wikipedia and statistical reports, featuring multi-level headers, cell merging, and hierarchical structure properties underrepresented in existing Table QA benchmarks.

We used HiTab's \textbf{test split} to evaluate two distinct tasks. 

First, for the Table Understanding Task, we use one original HiTab question--answer pair per table to evaluate how effectively different table representation formats convey complex table structures to LLMs. Each table is converted to six table representation formats (Markdown, CSV, HTML, LaTeX, MultiMarkdown, LTEM) while preserving original structure and content, with evaluation conducted using question--answer pairs for each format.

Second, as an additional task, we evaluate table generation performance from natural language inputs based on the same HiTab tables. Natural language generation requests are generated using GPT-5 based on each table's structure and values, with format-specific generation results analyzed through comparison with pre-generated reference tables for six table representation formats.

\subsection{Models and Evaluation Metrics}

\paragraph{Evaluation Models.}
To assess the consistency of format effects across model families, we conduct Table Understanding experiments with five LLMs and Table Generation experiments with three. Closed-weight models include GPT-5 and Gemini 2.5 Pro~\citep{openai2025gpt5systemcard,google2025gemini25report}. Open-weight models include the multimodal general-purpose Llama-4-Maverick-17B-128E-Instruct~\citep{meta2025llama4herd}, the large-parameter Qwen3-235B-A22B-Instruct-2507~\citep{yang2025qwen3}, and the reasoning-specialized DeepSeek-R1~\cite{deepseek2025r1}.

Task-specific prompt templates are held constant across models and formats, with format-dependent variation restricted to the table serialization or target output format and, where applicable, the corresponding syntax guidance. Temperature is set to each model's default value.

\paragraph{Table Understanding Evaluation.}
Table QA performance is primarily evaluated with normalized Exact Match (nEM), and we additionally report token-level F1 (tokenF1) to capture partial-match behavior~\citep{pasupat-liang-2015-compositional,cheng-etal-2022-hitab,rajpurkar2016squad}. Following numeric-answer evaluation practice in table QA~\citep{zhu-etal-2021-tat}, we compute numeric Exact Match (numEM) by extracting numbers and checking equality within a tolerance to reduce notation-induced discrepancies. To assess format stability, we decompose errors into Refusal (explicit answer avoidance) and Invalid (unnormalizable output or parsing failure); both are counted as incorrect, and their sum is reported as Total Error.

\paragraph{Table Generation Evaluation.}
For Table Generation, we prompt models to reconstruct HiTab tables \emph{from natural-language descriptions only} in a specified output format (Markdown/CSV/HTML/LaTeX/MultiMarkdown/LTEM). 
Unlike Table Understanding, \emph{exact format-level reproduction} is highly challenging, so we adopt \textbf{ANLS} (Average Normalized Levenshtein Similarity) as the primary metric to capture graded similarity between generated outputs and reference tables~\citep{biten2019scene}. 
We additionally report \textbf{Good+} (ANLS $\ge 0.7$) and \textbf{Excellent} (ANLS $\ge 0.9$) rates as practical success criteria; the full Excellent-rate breakdown is provided in Appendix~\ref{app:hitab_results}.

\paragraph{Token Efficiency Evaluation.}
Since input length varies by serialization format, we also report token-normalized metrics normalizing performance by input token count. For model $m$, format $f$, and benchmark/setting $s$, let input prompt token count be $Tok_{\text{prompt}}(m,f,s)$. When format-specific fixed-length grammar guides are included (e.g., LTEM), we separate this as a constant term:
\[
Tok_{\text{prompt}}^{\text{adj}}(m,f,s)=Tok_{\text{prompt}}(m,f,s)-g_f
\]
Token-normalized performance for Table Understanding is then:
\[
\mathrm{TNP}_{\mathrm{adj}}(m,f,s)=1000\times\frac{\mathrm{nEM}(m,f,s)}{Tok_{\text{prompt}}^{\text{adj}}(m,f,s)}
\]
This token/cost budget perspective for efficiency comparison is related to prior efficiency analyses \citep{han-etal-2025-token,zhang-etal-2025-lightthinker,zhang2024treacle,jung2023pcrl}.

\paragraph{Statistical Validation.}
For Table Understanding (QA), we performed three repeated runs per setting (model $\times$ format) on identical instances. We applied McNemar's test~\citep{mcnemar1947note} on paired predictions (Fisher-combined across runs) and report mean performance over runs, interpreting only consistent differences as significant.

\section{Results}

\subsection{Table Understanding}

\begin{figure}[t]
    \centering
    \includegraphics[width=0.93\linewidth]{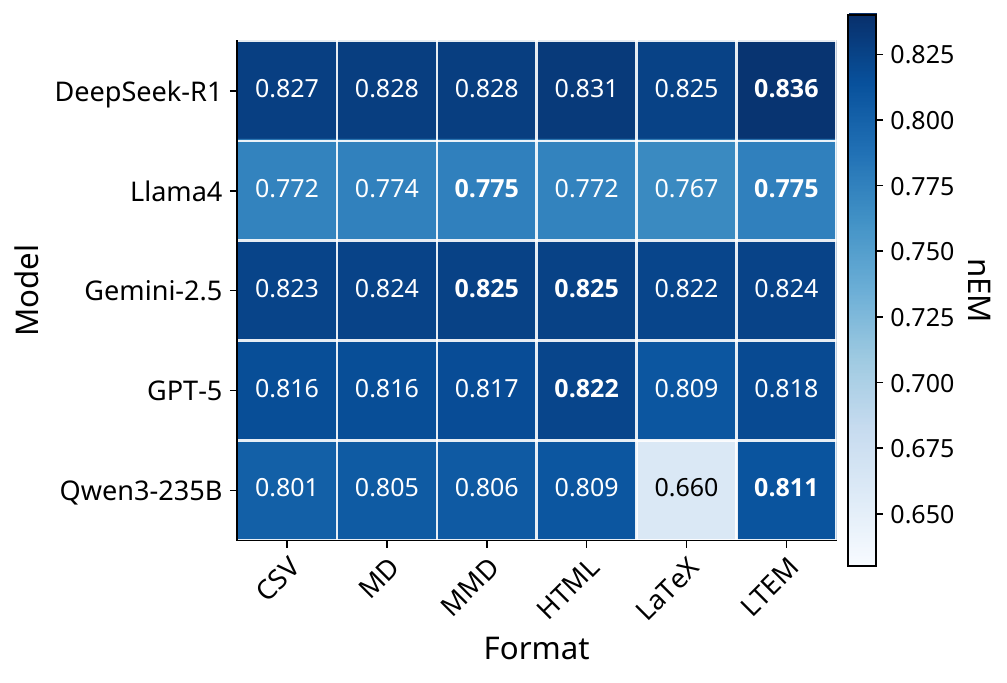}
    \caption{Model $\times$ serialization-format performance (nEM) on STB. Format sensitivity varies by model; notably, Qwen3-235B exhibits a large degradation under LaTeX serialization.}
    \label{fig:stb_model_format_heatmap}
\end{figure}

In this section, we analyze how table serialization formats affect Table QA accuracy and response stability. We first examine format effects on STB under controlled structural conditions, with each table--question instance serialized in all six formats. We then assess whether the observed trends extend to HiTab, a real-world hierarchical-table benchmark with a substantially different distribution of table sizes and structural properties.

On STB, LTEM achieves nEM comparable to HTML (0.813 vs.\ 0.812), while LaTeX is lower at 0.777 (Table~\ref{tab:stb_overall}). The gap is clearer in the error breakdown: LTEM keeps Refusal at 0.15\% and Invalid at 0.14\%, yielding the lowest Total Error, whereas LaTeX shows a sharp increase in Invalid to 2.67\% (Table~\ref{tab:stb_error_analysis}). Figure~\ref{fig:stb_model_format_heatmap} further shows that some models are particularly sensitive to LaTeX serialization, whereas LTEM remains stable under the same evaluation setting. Relative to HTML and LaTeX, LTEM provides a favorable accuracy--token trade-off (Table~\ref{tab:stb_overall}). McNemar tests show that LTEM significantly outperforms LaTeX for four of five models (Fisher-combined $p<0.01$); Gemini-2.5-Pro shows no significant difference (Appendix~\ref{app:mcnemar}). Failure-mode and question-type breakdowns are in Appendix~\ref{app:failure_qtype}.

\begin{figure}[!t]
\centering
\includegraphics[width=0.97\linewidth]{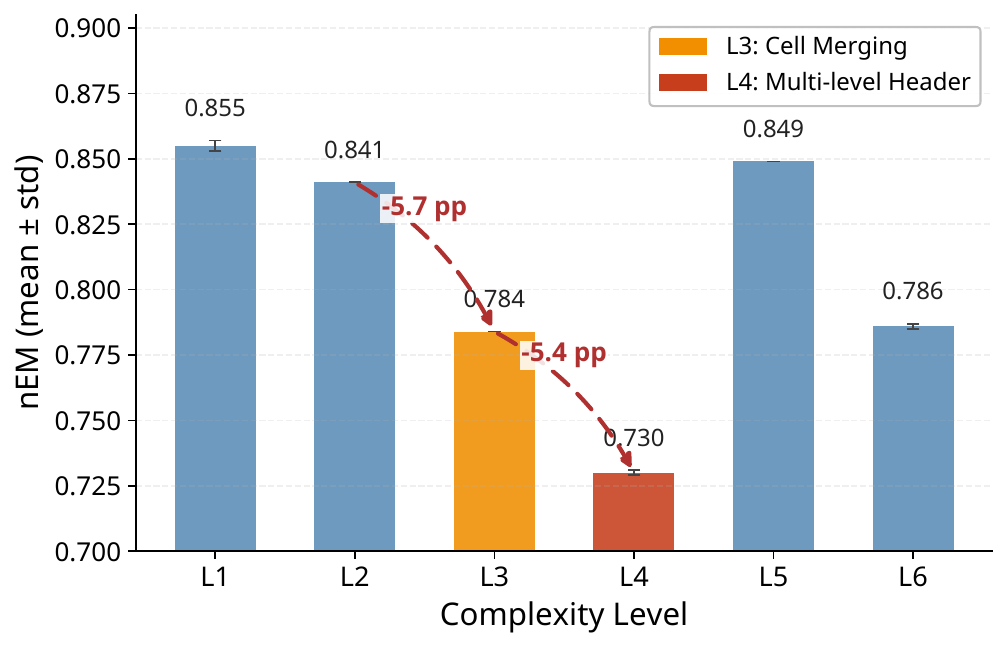}
\caption{Average performance (nEM) by structural complexity level (L1--L6) on STB. Arrow labels report absolute nEM changes in percentage points (pp). The highlighted L2--L3 and L3--L4 transitions correspond to the introduction of cell merging and multi-level headers.}
\label{fig:stb_complexity_levels}
\end{figure}

When broken down by Complexity Level, performance declines from L2 to L3 and further at L4 (Figure~\ref{fig:stb_complexity_levels}). These transitions reflect the point where models must reliably track span ranges and header roles as cell merging and multi-level headers appear. LTEM is designed to reduce ambiguity at this boundary by explicitly encoding spans and header nesting. Structure-conditioned results are reported in Appendix~\ref{app:structure_conditioned}. Additional level- and feature-wise breakdowns are in Appendix~\ref{app:fine_grained_understanding}.

\begin{figure}[!t]
\centering
\includegraphics[width=0.97\linewidth]{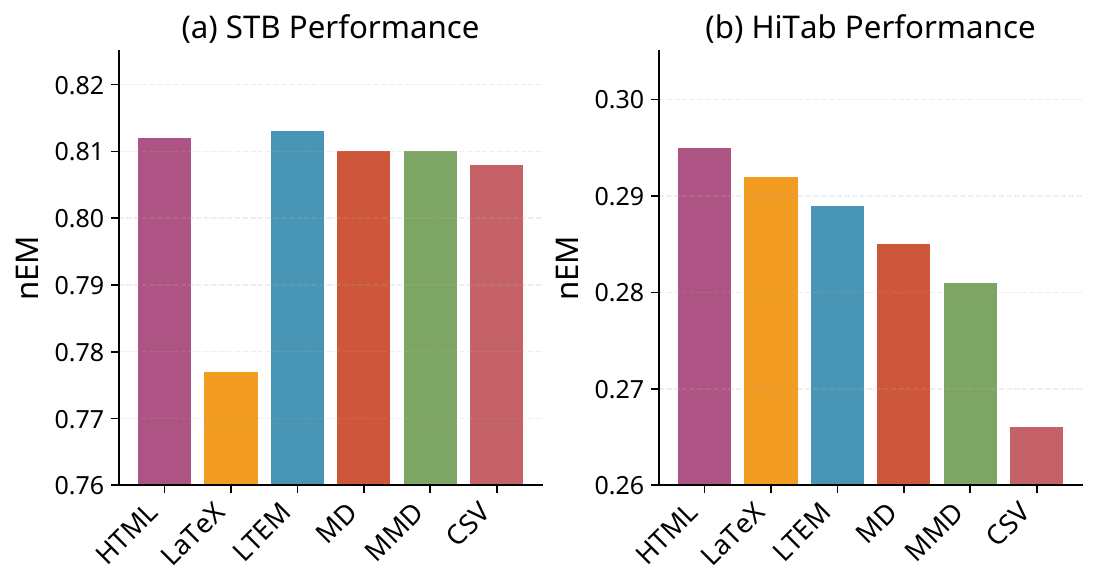}
\caption{Format-wise performance comparison (nEM) between STB and HiTab. The relative ordering of formats changes across datasets.}
\label{fig:stb_hitab_comparison}
\end{figure}

On HiTab, absolute accuracy decreases across formats and rankings become more dependent on dataset characteristics (Figure~\ref{fig:stb_hitab_comparison}). Average nEM is 0.295 for HTML, 0.292 for LaTeX, and 0.289 for LTEM, with a small 0.6 percentage-point gap (full results in Appendix~\ref{app:hitab_results}). Overall, on STB, LTEM achieves accuracy comparable to HTML and has a substantially lower Invalid rate than LaTeX (Tables~\ref{tab:stb_overall} and~\ref{tab:stb_error_analysis}).

\begin{table*}[!t]
\centering
\footnotesize
\setlength{\tabcolsep}{4pt}
\begin{tabular}{lcccccc}
\toprule
\textbf{Format} & \textbf{nEM$\uparrow$} & \textbf{TokenF1$\uparrow$} & \textbf{NumEM$\uparrow$} & \textbf{PromptTok$\downarrow$} & \textbf{TableTok$\downarrow$} & \textbf{TNP$_{\text{adj}}\uparrow$} \\
\midrule
\textbf{LTEM}   & \textbf{0.813\std{0.002}} & 0.844\std{0.002} & 0.943\std{0.000} & 449 & 228 & 1.809 \\
HTML            & \textbf{0.812\std{0.001}} & 0.844\std{0.000} & 0.945\std{0.000} & 612 & 392 & 1.326 \\
MMD             & 0.810\std{0.000} & 0.843\std{0.000} & 0.946\std{0.001} & 436 & 215 & 1.858 \\
MD              & 0.810\std{0.000} & 0.842\std{0.000} & 0.945\std{0.001} & 437 & 217 & 1.852 \\
CSV             & 0.808\std{0.001} & 0.840\std{0.001} & 0.940\std{0.000} & 370 & 149 & \textbf{2.183} \\
LaTeX           & 0.777\std{0.001} & 0.812\std{0.001} & 0.933\std{0.001} & 447 & 227 & 1.737 \\
\bottomrule
\end{tabular}
\caption{Overall STB performance by format (average over 5 models, 3 runs). PromptTok excludes LTEM guide tokens; TNP$_{\text{adj}}$\,=\,nEM/PromptTok$\times$1000. LTEM and HTML achieve comparable nEM. Shorter formats obtain higher token-normalized scores, while LTEM provides a favorable accuracy--token trade-off relative to HTML and LaTeX.}
\label{tab:stb_overall}
\end{table*}

\begin{table}[t]
\centering
\footnotesize
\renewcommand{\arraystretch}{0.95}
\setlength{\abovecaptionskip}{3pt}
\setlength{\belowcaptionskip}{0pt}
\setlength{\tabcolsep}{4pt}
\begin{tabular}{lccc}
\toprule
\textbf{Format} & \textbf{Refusal (\%)} & \textbf{Invalid (\%)} & \textbf{Total Error (\%)} \\
\midrule
\textbf{LTEM}   & \textbf{0.15} & 0.14 & 0.29 \\
CSV             & 0.22 & 0.09 & 0.31 \\
HTML            & 0.30 & 0.11 & 0.41 \\
MMD             & 0.34 & 0.15 & 0.49 \\
MD              & 0.46 & 0.15 & 0.61 \\
\textbf{LaTeX}  & 0.38 & \textbf{2.67} & \textbf{3.04} \\
\bottomrule
\end{tabular}
\caption{Error rate analysis on STB.
LTEM shows the lowest refusal rate.
    LaTeX exhibits a markedly higher invalid response rate than the other formats, primarily due to formatting and parsing failures.}
\label{tab:stb_error_analysis}
\end{table}

\subsection{Table Generation}
\label{sec:table_generation}

For the \textbf{Table Generation} task, we use HiTab tables as targets and evaluate whether a model can reconstruct the same tables \textit{from natural-language descriptions only} in a specified output format. Generation quality is evaluated using \textbf{ANLS} (normalized string similarity) and the \textbf{Good+} rate (ANLS $\ge 0.7$), while \textbf{PromptTok} denotes the token count of the user prompt.

As shown in Table~\ref{tab:gen_results_compact}, HTML achieves the best generation quality (ANLS 0.913) and the highest Good+ rate (96.2\%). LaTeX, MultiMarkdown, and LTEM form a similar middle-performing group, while Markdown and CSV perform worse as they cannot directly express complex structures (e.g., merged cells and hierarchical headers).

LTEM is designed to encode complex structures (F3--F4) rather than maximize generation scores; limited pretraining exposure constrains short-term performance, yet it remains close to MultiMarkdown (ANLS 0.720 vs.\ 0.734), suggesting it is a viable alternative in structure-heavy pipelines.

\begin{table}[htb]
\centering
\footnotesize
\renewcommand{\arraystretch}{0.95}
\setlength{\abovecaptionskip}{3pt}
\setlength{\belowcaptionskip}{0pt}
\setlength{\tabcolsep}{4pt}
\begin{tabular}{lccc}
\toprule
\textbf{Format} & \textbf{ANLS}$\uparrow$ & \textbf{Good+ (\%)} & \textbf{PromptTok}$\downarrow$ \\
\midrule
HTML            & \textbf{0.913} & \textbf{96.2} & 1149 \\
LaTeX          & 0.749 & 72.2 & 1168 \\
MultiMarkdown   & 0.734 & 69.0 & 1170 \\
\textbf{LTEM}   & 0.720 & 67.9 & 1429 \\
CSV             & 0.618 & 50.0 & 1170 \\
Markdown        & 0.549 & 41.2 & 1168 \\
\bottomrule
\end{tabular}
\caption{Table Generation performance (HiTab average; ANLS: normalized string similarity). PromptTok includes format-specific generation instructions, with an additional syntax guide for LTEM.}
\label{tab:gen_results_compact}
\end{table}

\section{Conclusion}
This paper introduced a \textbf{structural complexity taxonomy} (F1--F6, L1--L6) and \textbf{STB}, a controlled benchmark for analyzing how serialization formats affect LLM performance across complexity levels. We then proposed \textbf{LTEM}, a minimal Markdown extension that explicitly encodes cell spans and hierarchical headers, instantiating taxonomy-derived requirements as concrete syntax choices. Experiments on STB and HiTab show that LTEM achieves competitive table understanding accuracy while improving response stability at structural transitions where lightweight formats fail and verbose formats introduce noise. Beyond LTEM itself, we hope the taxonomy and benchmark provide reusable infrastructure for future research on robust and efficient LLM--table interaction.

\section*{Limitations}
\begin{itemize}
    \item \textbf{Dependence on data distribution.}
    Because the ranking of formats differs between STB (synthetic) and HiTab (real-world), it is difficult to generalize the superiority of a particular serialization format from a single benchmark. In practice, formats should be re-evaluated on a validation set whose distribution matches the target domain.

    \item \textbf{Training Distribution Bias.} 
    LLM pretraining corpora contain substantially more HTML and Markdown tables than LTEM-formatted content, as LTEM is introduced in this work. This exposure imbalance may favor established formats, particularly in generation tasks that depend on learned serialization patterns. HTML's strong generation performance (ANLS: 0.913, 76\% Excellent) may partly reflect this imbalance. In a separate table-QA setting, Appendix~\ref{app:fine_tuning_pilot} reports a small LoRA pilot showing that 4B-scale models can adapt to LTEM with limited supervision.

    \item \textbf{Model--format interactions.}
    Format effects vary substantially across models, and some model--format pairs exhibit atypical behaviors such as elevated invalid-response rates. Therefore, format selection cannot be made independently of the model, and prior validation on the deployment backbone is necessary.

    \item \textbf{Sensitivity to experimental settings (prompting/decoding).}
    Our experiments are based on a specific prompt template and decoding configuration. In particular, output length constraints, instruction wording, and post-processing rules can affect invalid/refusal rates for verbose formats such as LaTeX, and relative rankings may change under different settings.

    \item \textbf{Limited coverage of formats and domains.}
    While we compare six representative formats, the space of possible serialization designs is broader. Moreover, HiTab focuses primarily on Wikipedia tables, so our conclusions may not directly transfer to other industrial domains (e.g., OCR-extracted tables or noisy financial/log tables) without additional verification.
\end{itemize}

\section*{Acknowledgments}
This work was supported by the National Research Foundation of Korea (NRF) grant funded by the Korean government (MSIT) (No. RS-2025-0055621731482092640101); the Institute of Information \& Communications Technology Planning \& Evaluation (IITP) grant funded by the Korean government (MSIT) (RS-2019-II190004, \textit{Development of Semi-supervised Learning Language Intelligence Technology and Korean Tutoring Service for Foreigners}); a grant (26212MFDS008) from the Ministry of Food and Drug Safety in 2026; the Institute of Information \& Communications Technology Planning \& Evaluation (IITP) grant funded by the Korean government (MSIT) (RS-2026-25569031, \textit{Development of an AIBOM-Based Platform for Strengthening the Reliability and Security of the AI Supply Chain}); and the BK21 FOUR Program by Chungnam National University Research Grant, 2026.

\bibliography{custom}

\appendix

\onecolumn

\section{Complexity Level Definitions}
\label{app:complexity-levels}

\begin{table}[H]
\centering
\small
\resizebox{\columnwidth}{!}{%
\begin{tabular}{clp{4.5cm}p{3cm}}
\toprule
Level & Required Features & Description & Example \\
\midrule
\textbf{L1} & F1 & Atomic Grid & Simple data listing \\
\textbf{L2} & F1 + F2 & Headered Table & Standard table with headers \\
\textbf{L3} & F1 + F2 + F3 & Spanning Table & Table requiring cell merging \\
\textbf{L4} & F1 + F2 + F3 + F4 & Multi-level Header & Hierarchical header structure \\
\textbf{L5} & F1 + F2 + F3 + F4 + F5 & Block-partitioned & Same-schema section division \\
\textbf{L6} & F1 + F2 + F3 + F4 + F5 + F6 & Composite Table & Multi-schema combination \\
\bottomrule
\end{tabular}%
}
\caption{Complexity Level Definitions.}
\label{tab:complexity_levels}
\end{table}

\section{Additional Statistics of STB}
\subsection{Complexity and Domain Distributions of STB}
Tables~\ref{tab:stb-level} and~\ref{tab:stb-domain} summarize the complexity-level and domain distributions of STB, respectively.

\begin{table}[H]
\centering
\small
\setlength{\tabcolsep}{6pt}
\begin{tabular}{lrrrr}
\toprule
\textbf{Level} & \textbf{\#Tables} & \textbf{Share (\%)} & \textbf{Avg. rows} & \textbf{Avg. cols} \\
\midrule
L1 (Grid Only)         & 96  & 8.82  & 4.33  & 3.15 \\
L2 (Basic Header)      & 334 & 30.67 & 5.63  & 3.34 \\
L3 (Spanned)           & 342 & 31.40 & 7.15  & 5.22 \\
L4 (Multi-level Header)& 164 & 15.06 & 8.84  & 6.65 \\
L5 (Section Stacking)  & 87  & 7.99  & 12.66 & 4.20 \\
L6 (Composite)         & 66  & 6.06  & 15.55 & 11.50 \\
\midrule
\textbf{Total}         & 1{,}089 & 100.00 & -- & -- \\
\bottomrule
\end{tabular}
\caption{Complexity-level distribution and average table size of STB (tables containing all four intents: LOOKUP/COUNTING/COMPARISON/AGGREGATION). Average rows/cols aggregate all sub-tables for multi-table cases.}
\label{tab:stb-level}
\end{table}

\begin{table}[H]
\centering
\small
\setlength{\tabcolsep}{8pt}
\begin{tabular}{lrr}
\toprule
\textbf{Domain} & \textbf{\#Tables} & \textbf{Share (\%)} \\
\midrule
Scientific & 366 & 33.61 \\
Financial  & 286 & 26.26 \\
Survey     & 179 & 16.44 \\
Product    & 171 & 15.70 \\
Schedule   & 87  & 7.99 \\
\midrule
\textbf{Total} & 1{,}089 & 100.00 \\
\bottomrule
\end{tabular}
\caption{Domain distribution of STB (tables containing all four intents).}
\label{tab:stb-domain}
\end{table}

\section{LTEM Rendering Examples}
\label{app:rendering}

The syntax figures in Section~\ref{sec:ltem-core-syntax} show LTEM source only. Figure~\ref{fig:app-render-syntax} pairs each span construct with the table it renders to, and Figure~\ref{fig:app-render-report} shows a report table that combines them. Span markers are highlighted in the source, and merged cells are shaded in the rendered tables.

\begin{figure}[H]
\centering
\footnotesize
\renewcommand{\arraystretch}{1.1}
\setlength{\tabcolsep}{4pt}

\parbox{\textwidth}{\raggedright\textbf{(a) Column span (F3).} \texttt{<{}<} merges a cell into the cell on its left.}\\[3pt]
\begin{minipage}[c]{0.52\textwidth}
\begin{lstlisting}[style=ltemsrc]
| Time  | Room A   | Room B |
|-------|----------|--------|
| 09:00 | Keynote  | (*\ltemmk{<{}<}*)     |
| 10:00 | Workshop | Panel  |
\end{lstlisting}
\end{minipage}\hfill$\Rightarrow$\hfill
\begin{minipage}[c]{0.42\textwidth}\centering
\begin{tabular}{|l|l|l|}
\hline
\textbf{Time} & \textbf{Room A} & \textbf{Room B} \\ \hline
09:00 & \multicolumn{2}{l|}{\cellcolor{ltemmerge}Keynote} \\ \hline
10:00 & Workshop & Panel \\ \hline
\end{tabular}
\end{minipage}

\vspace{6pt}
\parbox{\textwidth}{\raggedright\textbf{(b) Row span (F3).} \texttt{\textasciicircum\textasciicircum} merges a cell into the cell above.}\\[3pt]
\begin{minipage}[c]{0.52\textwidth}
\begin{lstlisting}[style=ltemsrc]
| Region | City   | Sales |
|--------|--------|-------|
| Asia   | Seoul  | 120   |
| (*\ltemmk{\textasciicircum\textasciicircum}*)     | Tokyo  | 95    |
| Europe | Paris  | 80    |
| (*\ltemmk{\textasciicircum\textasciicircum}*)     | Berlin | 70    |
\end{lstlisting}
\end{minipage}\hfill$\Rightarrow$\hfill
\begin{minipage}[c]{0.42\textwidth}\centering
\begin{tabular}{|l|l|l|}
\hline
\textbf{Region} & \textbf{City} & \textbf{Sales} \\ \hline
\cellcolor{ltemmerge} & Seoul & 120 \\ \cline{2-3}
\cellcolor{ltemmerge}\multirow{-2}{*}{Asia} & Tokyo & 95 \\ \hline
\cellcolor{ltemmerge} & Paris & 80 \\ \cline{2-3}
\cellcolor{ltemmerge}\multirow{-2}{*}{Europe} & Berlin & 70 \\ \hline
\end{tabular}
\end{minipage}

\vspace{6pt}
\parbox{\textwidth}{\raggedright\textbf{(c) Span attribute (F3).} \texttt{\{c:5\}} spans five columns, the same as writing \texttt{<{}<} four times.}\\[3pt]
\begin{minipage}[c]{0.52\textwidth}
\begin{lstlisting}[style=ltemsrc]
| Time  | Mon  | Tue  | Wed  | Thu  | Fri  |
|-------|------|------|------|------|------|
| 07:00 | Yoga | Spin | Yoga | Spin | Yoga |
| 12:00 | Closed (*\ltemmk{\{c:5\}}*)                     |
| 18:00 | HIIT | Yoga | HIIT | Yoga | HIIT |
\end{lstlisting}
\end{minipage}\hfill$\Rightarrow$\hfill
\begin{minipage}[c]{0.42\textwidth}\centering
\begin{tabular}{|l|l|l|l|l|l|}
\hline
\textbf{Time} & \textbf{Mon} & \textbf{Tue} & \textbf{Wed} & \textbf{Thu} & \textbf{Fri} \\ \hline
07:00 & Yoga & Spin & Yoga & Spin & Yoga \\ \hline
12:00 & \multicolumn{5}{l|}{\cellcolor{ltemmerge}Closed} \\ \hline
18:00 & HIIT & Yoga & HIIT & Yoga & HIIT \\ \hline
\end{tabular}
\end{minipage}

\vspace{6pt}
\parbox{\textwidth}{\raggedright\textbf{(d) Block span (F3).} \texttt{\{c:2,r:2\}} spans two columns and two rows; the covered cells are left empty.}\\[3pt]
\begin{minipage}[c]{0.52\textwidth}
\begin{lstlisting}[style=ltemsrc]
| Time  | Room A             | Room B | Room C |
|-------|--------------------|--------|--------|
| 09:00 | Workshop (*\ltemmk{\{c:2,r:2\}}*) |        | Talk   |
| 10:00 |                    |        | Demo   |
| 11:00 | Standup            | Lunch  | Review |
\end{lstlisting}
\end{minipage}\hfill$\Rightarrow$\hfill
\begin{minipage}[c]{0.42\textwidth}\centering
\begin{tabular}{|l|l|l|l|}
\hline
\textbf{Time} & \textbf{Room A} & \textbf{Room B} & \textbf{Room C} \\ \hline
09:00 & \multicolumn{2}{l|}{\cellcolor{ltemmerge}} & Talk \\ \cline{1-1}\cline{4-4}
10:00 & \multicolumn{2}{l|}{\cellcolor{ltemmerge}\multirow{-2}{*}{Workshop}} & Demo \\ \hline
11:00 & Standup & Lunch & Review \\ \hline
\end{tabular}
\end{minipage}

\vspace{6pt}
\parbox{\textwidth}{\raggedright\textbf{(e) Multi-level header (F4).} Every row above the separator is a header row; \texttt{<{}<} and \texttt{\textasciicircum\textasciicircum} group the headers.}\\[3pt]
\begin{minipage}[c]{0.52\textwidth}
\begin{lstlisting}[style=ltemsrc]
| Region | Revenue | (*\ltemmk{<{}<}*)  |
| (*\ltemmk{\textasciicircum\textasciicircum}*)     | Q1      | Q2  |
|--------|---------|-----|
| Asia   | 100     | 120 |
| Europe | 90      | 110 |
\end{lstlisting}
\end{minipage}\hfill$\Rightarrow$\hfill
\begin{minipage}[c]{0.42\textwidth}\centering
\begin{tabular}{|l|l|l|}
\hline
\cellcolor{ltemmerge} & \multicolumn{2}{c|}{\cellcolor{ltemmerge}\textbf{Revenue}} \\ \cline{2-3}
\cellcolor{ltemmerge}\multirow{-2}{*}{\textbf{Region}} & \textbf{Q1} & \textbf{Q2} \\ \hline
Asia & 100 & 120 \\ \hline
Europe & 90 & 110 \\ \hline
\end{tabular}
\end{minipage}

\vspace{6pt}
\parbox{\textwidth}{\raggedright\textbf{(f) Section rows (F5).} A row merged across all columns starts a new section that shares the same columns.}\\[3pt]
\begin{minipage}[c]{0.52\textwidth}
\begin{lstlisting}[style=ltemsrc]
| Card            | Annual fee | Rewards  |
|-----------------|------------|----------|
| Travel cards    | (*\ltemmk{<{}<}*)         | (*\ltemmk{<{}<}*)       |
| Globe           | $95        | 2x miles |
| Atlas           | $250       | 3x miles |
| Cash-back cards | (*\ltemmk{<{}<}*)         | (*\ltemmk{<{}<}*)       |
| Daily           | $0         | 1.5%     |
\end{lstlisting}
\end{minipage}\hfill$\Rightarrow$\hfill
\begin{minipage}[c]{0.42\textwidth}\centering
\begin{tabular}{|l|l|l|}
\hline
\textbf{Card} & \textbf{Annual fee} & \textbf{Rewards} \\ \hline
\multicolumn{3}{|l|}{\cellcolor{ltemmerge}Travel cards} \\ \hline
Globe & \$95 & 2x miles \\ \hline
Atlas & \$250 & 3x miles \\ \hline
\multicolumn{3}{|l|}{\cellcolor{ltemmerge}Cash-back cards} \\ \hline
Daily & \$0 & 1.5\% \\ \hline
\end{tabular}
\end{minipage}

\caption{LTEM source (left) and the rendered table (right) for each span construct.}
\label{fig:app-render-syntax}
\end{figure}

\begin{figure}[H]
\centering
\footnotesize
\renewcommand{\arraystretch}{1.1}
\setlength{\tabcolsep}{4pt}
\begin{minipage}{0.86\textwidth}
\begin{lstlisting}[style=ltemsrc]
| Area        | Program              | FY2025 | (*\ltemmk{<{}<}*)    | FY2026 | (*\ltemmk{<{}<}*)    | Change |
| (*\ltemmk{\textasciicircum\textasciicircum}*)          | (*\ltemmk{\textasciicircum\textasciicircum}*)                   | Budget | Spent | Budget | Spent | (*\ltemmk{\textasciicircum\textasciicircum}*)     |
|-------------|----------------------|-------:|------:|-------:|------:|-------:|
| Digital     | E-document standards |  1,240 | 1,180 |  1,380 |   612 | +11.3% |
| (*\ltemmk{\textasciicircum\textasciicircum}*)          | Open public data     |    860 |   845 |    910 |   402 |  +5.8% |
| (*\ltemmk{\textasciicircum\textasciicircum}*)          | Data structuring     |      - |     - |    420 |    95 |      - |
| Regional    | Regional hubs        |  2,100 | 2,064 |  2,050 | 1,010 |  -2.4% |
| (*\ltemmk{\textasciicircum\textasciicircum}*)          | Youth settlement     |    730 |   701 |    790 |   355 |  +8.2% |
| Grand total | (*\ltemmk{<{}<}*)                   |  4,930 | 4,790 |  5,550 | 2,474 | +12.6% |
\end{lstlisting}
\end{minipage}

\vspace{2pt}
$\Downarrow$
\vspace{2pt}

\begin{tabular}{|l|l|r|r|r|r|r|}
\hline
\cellcolor{ltemmerge} & \cellcolor{ltemmerge} & \multicolumn{2}{c|}{\cellcolor{ltemmerge}\textbf{FY2025}} & \multicolumn{2}{c|}{\cellcolor{ltemmerge}\textbf{FY2026}} & \cellcolor{ltemmerge} \\ \cline{3-6}
\cellcolor{ltemmerge}\multirow{-2}{*}{\textbf{Area}} & \cellcolor{ltemmerge}\multirow{-2}{*}{\textbf{Program}} & \textbf{Budget} & \textbf{Spent} & \textbf{Budget} & \textbf{Spent} & \cellcolor{ltemmerge}\multirow{-2}{*}{\textbf{Change}} \\ \hline
\cellcolor{ltemmerge} & E-document standards & 1,240 & 1,180 & 1,380 & 612 & +11.3\% \\ \cline{2-7}
\cellcolor{ltemmerge} & Open public data & 860 & 845 & 910 & 402 & +5.8\% \\ \cline{2-7}
\cellcolor{ltemmerge}\multirow{-3}{*}{Digital} & Data structuring & - & - & 420 & 95 & - \\ \hline
\cellcolor{ltemmerge} & Regional hubs & 2,100 & 2,064 & 2,050 & 1,010 & -2.4\% \\ \cline{2-7}
\cellcolor{ltemmerge}\multirow{-2}{*}{Regional} & Youth settlement & 730 & 701 & 790 & 355 & +8.2\% \\ \hline
\multicolumn{2}{|l|}{\cellcolor{ltemmerge}Grand total} & 4,930 & 4,790 & 5,550 & 2,474 & +12.6\% \\ \hline
\end{tabular}

\caption{A report table that combines row spans, column spans, and a two-row header (F3--F4), as LTEM source (top) and the rendered table (bottom). Numeric columns use the Markdown right-alignment marker (\texttt{-{}-{}-:}).}
\label{fig:app-render-report}
\end{figure}

\section{Formal Grammar (EBNF)}
\label{app:ltem-ebnf}

This appendix defines the \textbf{syntax} of LTEM using Extended Backus--Naur Form (EBNF).
The grammar is a context-free grammar (CFG) and follows the notation of ISO/IEC 14977:1996.

\subsection{Notation Conventions}

\begin{table}[H]
\centering
\begin{tabular}{ll}
\toprule
\textbf{Symbol} & \textbf{Meaning} \\
\midrule
\texttt{=} & Definition \\
\texttt{,} & Concatenation \\
\texttt{|} & Alternation \\
\texttt{[ ]} & Optional (0 or 1 occurrence) \\
\texttt{\{ \}} & Repetition (0 or more occurrences) \\
\texttt{( )} & Grouping \\
\texttt{" "} & Terminal string \\
\texttt{'...'} & Terminal string (alternative) \\
\texttt{? ?} & Special sequence (e.g., regular expression) \\
\texttt{(* *)} & Comment \\
\bottomrule
\end{tabular}
\caption{Notation conventions used in the EBNF specification.}
\label{tab:ltem-ebnf-notation}
\end{table}

\subsection{Document Structure}

\begin{lstlisting}[language=ebnf]
(* LTEM Document - Top Level *)
document = { table } ;

(* Table Structure *)
table = [ table_id_row ], header_section, separator_row, [ data_section ] ;

table_id_row = "|", "#", whitespace, table_id, whitespace, "|", newline ;

table_id = letter_or_underscore, { letter_or_digit_or_underscore } ;

header_section = row, { row } ;

data_section = row, { row } ;

separator_row = "|", separator_cell, { "|", separator_cell }, "|", newline ;

separator_cell = { "-" }, [ ":" ], { "-" } ;

row = "|", cell, { "|", cell }, "|", [ newline ] ;
\end{lstlisting}

\subsection{Cell Structure}

\begin{lstlisting}[language=ebnf]
(* Cell Definition *)
cell = cell_content, [ span_attribute ]
     | colspan_marker
     | rowspan_marker
     | empty_cell ;

empty_cell = { whitespace } ;

cell_content = { content_element } ;

content_element = plain_text
                | table_reference
                | inline_math
                | display_math
                | inline_list
                | line_break
                | checkbox ;

plain_text = ? any character except |, {, }, \, $, @, newline ? ;
\end{lstlisting}

\subsection{Span Syntax}

\begin{lstlisting}[language=ebnf]
(* Span Markers *)
colspan_marker = { whitespace }, "<<", { whitespace } ;

rowspan_marker = { whitespace }, "^^", { whitespace } ;

(* Span Attributes *)
span_attribute = "{", span_spec, { ",", span_spec }, "}" ;

span_spec = colspan_spec | rowspan_spec ;

colspan_spec = "c", ":", positive_integer ;

rowspan_spec = "r", ":", positive_integer ;

positive_integer = nonzero_digit, { digit } ;
\end{lstlisting}

\subsection{Reference Syntax (Phase 4)}

\begin{lstlisting}[language=ebnf]
(* Table References *)
table_reference = simple_reference
                | filtered_reference
                | column_reference
                | cell_reference ;

simple_reference = "@", table_id ;

filtered_reference = "@", table_id, "[", filter_condition, "]" ;

column_reference = "@", table_id, ".", column_name ;

cell_reference = "@", table_id, "[", row_index, "]", ".", column_name ;

filter_condition = column_name, comparison_op, value ;

comparison_op = "<" | ">" | "<=" | ">=" | "=" | "!=" ;

column_name = letter_or_underscore, { letter_or_digit_or_underscore } ;

row_index = digit, { digit } ;

value = number | string_literal ;

number = [ "-" ], digit, { digit }, [ ".", digit, { digit } ] ;

string_literal = "'", { ? any character except ' ? }, "'" ;
\end{lstlisting}

\subsection{Inline Content Syntax}

\begin{lstlisting}[language=ebnf]
(* Line Break *)
line_break = "\\", "\\" ;

(* Inline Math *)
inline_math = "$", math_content, "$" ;

display_math = "$$", math_content, "$$" ;

math_content = { ? any character except $ ? } ;

(* Inline Lists *)
inline_list = list_item, { line_break, list_item } ;

list_item = bullet_item | numbered_item | checkbox_item ;

bullet_item = "-", whitespace, item_text ;

numbered_item = digit, { digit }, ".", whitespace, item_text ;

checkbox_item = checkbox, whitespace, item_text ;

checkbox = "[", ( "x" | " " ), "]" ;

item_text = { ? any character except \\, |, newline ? } ;
\end{lstlisting}

\subsection{Terminal Definitions}

\begin{lstlisting}[language=ebnf]
(* Character Classes *)
letter = "A" | "B" | ... | "Z" | "a" | "b" | ... | "z" ;

digit = "0" | "1" | "2" | "3" | "4" | "5" | "6" | "7" | "8" | "9" ;

nonzero_digit = "1" | "2" | "3" | "4" | "5" | "6" | "7" | "8" | "9" ;

letter_or_underscore = letter | "_" ;

letter_or_digit_or_underscore = letter | digit | "_" ;

whitespace = " " | "\t" ;

newline = "\n" | "\r\n" ;
\end{lstlisting}

\subsection{Grammar Properties}

\begin{table}[H]
\centering
\begin{tabular}{lll}
\toprule
\textbf{Property} & \textbf{Value} & \textbf{Implication} \\
\midrule
Chomsky Type & Type 2 (CFG) & Efficient parsing (O(n\textsuperscript{3}) worst, O(n) practical) \\
Ambiguity & Unambiguous & Deterministic parsing \\
Left Recursion & None & LL parser compatible \\
Lookahead & LL(1) for most constructs & Simple recursive descent \\
Unicode & Supported in content & UTF-8 encoding \\
\bottomrule
\end{tabular}
\caption{Summary of grammar properties.}
\label{tab:ltem-ebnf-properties}
\end{table}

\subsection{Lark Implementation}

The following is a Lark parser grammar derived from the EBNF above:

\begin{lstlisting}[language=]
// LTEM Grammar for Lark Parser
// Version 1.0

start: document

document: table+

table: table_id_row? header_section separator_row data_section?

table_id_row: "|" "#" TABLE_ID "|" NEWLINE

header_section: row+

data_section: row+

separator_row: "|" SEP_CELL ("|" SEP_CELL)* "|" NEWLINE
SEP_CELL: /:?-+:?/

row: "|" cell ("|" cell)* "|" NEWLINE?

cell: cell_content span_attr?
    | COLSPAN_MARKER
    | ROWSPAN_MARKER
    | EMPTY_CELL

cell_content: content_element+

content_element: TEXT
               | table_ref
               | INLINE_MATH
               | DISPLAY_MATH
               | LINE_BREAK
               | CHECKBOX

span_attr: "{" span_spec ("," span_spec)* "}"
span_spec: COLSPAN_ATTR | ROWSPAN_ATTR

COLSPAN_ATTR: "c:" POSITIVE_INT
ROWSPAN_ATTR: "r:" POSITIVE_INT
POSITIVE_INT: /[1-9][0-9]*/

COLSPAN_MARKER: /\s*<<\s*/
ROWSPAN_MARKER: /\s*\^\^\s*/
EMPTY_CELL: /\s*/

table_ref: REF_CELL | REF_COLUMN | REF_FILTERED | REF_SIMPLE
REF_SIMPLE: /@[a-zA-Z_][a-zA-Z0-9_]*/
REF_FILTERED: /@[a-zA-Z_][a-zA-Z0-9_]*\[[^\]]+\]/
REF_COLUMN: /@[a-zA-Z_][a-zA-Z0-9_]*\.[a-zA-Z_][a-zA-Z0-9_]*/
REF_CELL: /@[a-zA-Z_][a-zA-Z0-9_]*\[[0-9]+\]\.[a-zA-Z_][a-zA-Z0-9_]*/

INLINE_MATH: /\$[^$]+\$/
DISPLAY_MATH: /\$\$[^$]+\$\$/

LINE_BREAK: /\s*\\\\\s*/

CHECKBOX: /\[[ x]\]/

TABLE_ID: /[a-zA-Z_][a-zA-Z0-9_]*/
TEXT: /[^|{}\n\\$@]+/
NEWLINE: /\r?\n/

%import common.WS_INLINE
%ignore WS_INLINE
\end{lstlisting}

\section{Prompt Templates}
\label{app:prompts}

This appendix reports the prompt templates used for STB QA generation, QA verification, and table-understanding evaluation. Curly-brace fields denote runtime variables filled by the pipeline.

\begin{tcolorbox}[breakable,colback=orange!5!white,colframe=orange!80!black,title=STB QA Generation Prompt]
\begin{lstlisting}[basicstyle=\ttfamily\footnotesize,breaklines=true,frame=none]
Create 1 table QA as JSON.
INTENT={intent}
Return ONLY: question, answer, targets. {targets_rule}
Rules: answerable ONLY from Cells; question must NOT mention [header,rowspan,colspan,merge,layout,structure,level,index,row,column]; answer must be short (number/short phrase), no explanation; Targets should list only data cells that directly contribute to the final answer. Compute the answer considering the units and symbols shown in the table.
Format: table_key(optional)|row,col,rowspan,colspan,H|D|text (H=header, D=data). {target_hint}
{intent_guidance}
{level_hint}
{multi_hint}
Cells:
{table_lines}
Output exactly: {output_schema}
\end{lstlisting}
\end{tcolorbox}

The intent guidance field is instantiated as follows: LOOKUP reads one value for a specific item and attribute; AGGREGATION computes a numeric summary over multiple numeric cells; COUNTING returns an integer count of items meeting a condition; and COMPARISON compares multiple numeric values and selects the higher, lower, best, or worst item.

\begin{tcolorbox}[breakable,colback=orange!5!white,colframe=orange!80!black,title=QA Verification Prompt]
\begin{lstlisting}[basicstyle=\ttfamily\footnotesize,breaklines=true,frame=none]
You are a strict validator.
Answer the question using ONLY the table cells.
Return JSON ONLY: {"answer":"..."}.
No explanation. No extra keys. Answer must be short.

{fmt_line}

Question: {question}
Cells:
{table_lines}
\end{lstlisting}
\end{tcolorbox}

\begin{tcolorbox}[breakable,colback=orange!5!white,colframe=orange!80!black,title=Table Understanding Evaluation Prompt]
\begin{lstlisting}[basicstyle=\ttfamily\footnotesize,breaklines=true,frame=none]
You are an expert data analyst answering questions from tables. Use only the table; if answer missing, output N/A. Respond with ONLY the final answer value (no explanation, no prefix). If a number has commas, omit commas. Include units only if shown in the table.
Table format: {fmt}.{fmt_hint}

TABLE:
{table_text}

QUESTION: {question}
ANSWER :
\end{lstlisting}
\end{tcolorbox}

For LTEM inputs, the evaluation prompt additionally includes the following guide:

\begin{tcolorbox}[breakable,colback=orange!5!white,colframe=orange!80!black,title=LTEM Guide Added to Evaluation Prompt]
\begin{lstlisting}[basicstyle=\ttfamily\footnotesize,breaklines=true,frame=none]
LTEM guide: '<<' = copy nearest non-<< cell to the left; '^^' = copy nearest non-^^ cell above; {c:N}/{r:N} = explicit col/row spans; '|---|' separates headers; '\\' is a newline. Resolve merges/spans first and answer from the resolved table.
\end{lstlisting}
\end{tcolorbox}

All token counts reported in the paper are computed from the stored prompts using the \texttt{o200k\_base} tokenizer.

\section{STB Generation Pipeline}
\label{app:stb_pipeline}

STB generation follows a controlled three-stage pipeline. In the first stage, a rule-based structural generator samples the complexity level, domain, row/column count, and the structural operations (F1--F6) required for each table. In the second stage, large language models populate each table with domain-appropriate headers and cell values while preserving the sampled structure. In the third stage, question--answer pairs are generated for four intent types (LOOKUP, COUNTING, COMPARISON, AGGREGATION) and filtered by agreement between two independent model families. Prompt templates used in each stage are provided in Appendix~\ref{app:prompts}.

\section{HiTab Overall Results}
\label{app:hitab_results}

Per-format performance on the HiTab test split is shown below. The STB overall results appear as Table~\ref{tab:stb_overall} in the main body.

\begin{table}[H]
\centering
\small
\begin{tabular}{lcccccc}
\toprule
\textbf{Format} & \textbf{nEM$\uparrow$} & \textbf{TokenF1$\uparrow$} & \textbf{NumEM$\uparrow$} & \textbf{PromptTok$\downarrow$} & \textbf{TNP$_{\text{adj}}\uparrow$} & \textbf{Rank} \\
\midrule
\textbf{HTML}   & \textbf{0.295\std{0.001}} & 0.331\std{0.002} & 0.519\std{0.001} & 1310 & 0.225 & \textbf{1} \\
LaTeX           & 0.292\std{0.003} & 0.333\std{0.003} & 0.527\std{0.002} & 966  & 0.303 & 2 \\
LTEM            & 0.289\std{0.005} & 0.329\std{0.004} & 0.504\std{0.004} & 1101 & 0.262 & 3 \\
MD              & 0.285\std{0.003} & 0.322\std{0.004} & 0.515\std{0.003} & 1078 & 0.264 & 4 \\
MMD             & 0.281\std{0.001} & 0.320\std{0.001} & 0.520\std{0.002} & 1058 & 0.266 & 5 \\
CSV             & 0.266\std{0.004} & 0.306\std{0.003} & 0.515\std{0.003} & 770  & 0.345 & 6 \\
\bottomrule
\end{tabular}
\caption{Overall performance on the HiTab test split (515 hierarchical tables). HTML achieves the highest nEM, followed by LaTeX. The ranking of LaTeX reverses compared to STB (rank 6$\rightarrow$2), suggesting format effectiveness is dataset-dependent.}
\label{tab:hitab_overall}
\end{table}

\paragraph{Table Generation success rates.}
Table~\ref{tab:appendix_generation_excellent} reports the Excellent-rate breakdown corresponding to the Table Generation results in Table~\ref{tab:gen_results_compact}.

\begin{table}[H]
\centering
\small
\begin{tabular}{lr}
\toprule
\textbf{Format} & \textbf{Excellent (\%)} \\
\midrule
\textbf{HTML}   & \textbf{76.2} \\
LaTeX           & 29.3 \\
MultiMarkdown   & 35.4 \\
LTEM            & 25.8 \\
CSV             & 25.4 \\
Markdown        & 22.5 \\
\bottomrule
\end{tabular}
\caption{Excellent-rate breakdown for Table Generation on HiTab. Excellent denotes ANLS $\ge 0.9$; the values use the same evaluation setting as Table~\ref{tab:gen_results_compact}.}
\label{tab:appendix_generation_excellent}
\end{table}

\section{Supplementary Analyses}
\label{app:additional_analyses}

This appendix consolidates additional fine-grained analyses that characterize format behavior beyond the aggregate results in the main paper. Unless otherwise stated, the reported STB understanding numbers are normalized exact match (nEM) over the same evaluation protocol as the main paper. These results are intended to clarify where format differences arise rather than to replace the primary aggregate comparisons in the main Results section.

\subsection{Fine-grained STB Understanding Analyses}
\label{app:fine_grained_understanding}
\label{app:level_feature_breakdown}

\paragraph{Level- and feature-wise Markdown--LTEM comparison.}

Table~\ref{tab:appendix_levelwise_md_ltem} compares Markdown and LTEM across the STB complexity levels. The nEM differences are small, with LTEM slightly higher at each level. Invalid rates are low for both formats and do not show a uniform level-wise advantage. This supports a conservative interpretation: LTEM's main distinction is explicit structural representation rather than a uniform large QA gain.

\begin{table}[H]
\centering
\small
\begin{tabular}{lrrrrr}
\toprule
\textbf{Level} & \textbf{N} & \textbf{MD nEM} & \textbf{LTEM nEM} & \textbf{$\Delta$} & \textbf{Invalid MD/LTEM} \\
\midrule
L1 & 96  & 0.863 & 0.866 & +0.003 & 0.05/0.05 \\
L2 & 334 & 0.845 & 0.847 & +0.002 & 0.22/0.13 \\
L3 & 342 & 0.787 & 0.791 & +0.004 & 0.11/0.16 \\
L4 & 164 & 0.735 & 0.737 & +0.002 & 0.17/0.19 \\
L5 & 87  & 0.858 & 0.863 & +0.004 & 0.00/0.00 \\
L6 & 66  & 0.788 & 0.793 & +0.005 & 0.28/0.20 \\
\bottomrule
\end{tabular}
\caption{Level-wise comparison of Markdown and LTEM on STB. Invalid rates are percentages.}
\label{tab:appendix_levelwise_md_ltem}
\end{table}

The same pattern appears when conditioning on structural features (Table~\ref{tab:appendix_featurewise_md_ltem}). The main distinction starts at F3, cell merging: Markdown can only approximate such tables through visual alignment or repeated text, whereas LTEM explicitly marks span extent.

\begin{table}[H]
\centering
\small
\begin{tabular}{lrrrrr}
\toprule
\textbf{Feature slice} & \textbf{N} & \textbf{MD nEM} & \textbf{LTEM nEM} & \textbf{$\Delta$} & \textbf{Invalid MD/LTEM} \\
\midrule
F2 & 1{,}057 & 0.809 & 0.812 & +0.003 & 0.16/0.14 \\
F3 & 593  & 0.783 & 0.787 & +0.004 & 0.11/0.15 \\
F4 & 345  & 0.762 & 0.766 & +0.004 & 0.17/0.20 \\
F5 & 87 & 0.858 & 0.863 & +0.004 & 0.00/0.00 \\
F6 & 66   & 0.788 & 0.793 & +0.005 & 0.28/0.20 \\
\bottomrule
\end{tabular}
\caption{Feature-conditioned Markdown--LTEM comparison on STB. F2--F6 denote operational slices in which the corresponding structural feature is observed; the slices may overlap and are not cumulative level subsets. Invalid rates are percentages.}
\label{tab:appendix_featurewise_md_ltem}
\end{table}

\subsection{LTEM Design-Choice Ablation}
\label{app:ltem_design_ablation}

To examine the sensitivity of Table QA performance to LTEM's surface syntax, we compare the proposed forms of five elements with structure-preserving alternatives on matched STB instances. For each comparison, table content, questions, reference answers, and the evaluation protocol are held fixed, and only the notation of the target element is varied. Experiments use GPT-5, and paired differences are assessed using exact McNemar tests.

\begin{table}[H]
\centering
\small
\setlength{\tabcolsep}{3pt}
\begin{tabular}{lccrr}
\toprule
\textbf{Element ($N$)} & \textbf{LTEM} & \textbf{Alternative} & \textbf{nEM (\%)} & \textbf{\textit{p}} \\
\midrule
Colspan (400)      & \texttt{<<} & \texttt{\{c:N\}} & 77.8 / 78.8 & .481 \\
Rowspan (348)      & \texttt{\textasciicircum\textasciicircum} & \texttt{\{r:N\}} & 75.6 / 76.7 & .454 \\
Header scope (288) & Separator & Header count & 75.3 / 77.1 & .302 \\
Section label (348)& Marker row & Attribute row & 85.6 / 85.9 & 1.000 \\
Table ID (264)     & In-grid row & MD heading & 83.3 / 82.2 & .453 \\
\bottomrule
\end{tabular}
\caption{Element-wise LTEM design-choice ablation on STB using GPT-5. Each row compares the LTEM form with a structure-preserving alternative on the same QA pairs ($N$). nEM is reported as LTEM/alternative; \textit{p}-values are from exact McNemar tests.}
\label{tab:appendix_ltem_design_ablation}
\end{table}

\Needspace{4\baselineskip}
None of the five paired comparisons reaches statistical significance (all $p \ge .302$); the ablation therefore does not identify a preferred surface form on the basis of QA accuracy. LTEM adopts the proposed forms in accordance with the Markdown Compatibility and Linear Readability principles.

\subsection{Structure-conditioned Accuracy}
\label{app:structure_conditioned}

We stratify L1--L5 single-table examples by span-continuation and repeated-value ratios, using None/Low/Medium/High cutoffs of 0/.15/.35 and 0/.30/.60, respectively. Table~\ref{tab:appendix_structure_conditioned} reports descriptive nEM results. Across the larger slices, LTEM remains comparable to HTML and Markdown, while LaTeX is generally lower; the small High buckets are reported without broad generalization.

\begin{table}[H]
\centering
\small
\begin{tabular}{llrrrr}
\toprule
\textbf{Condition} & \textbf{Bucket} & \textbf{HTML} & \textbf{LaTeX} & \textbf{LTEM} & \textbf{MD} \\
\midrule
Merged density & None (430) & 0.852 & 0.811 & 0.852 & 0.849 \\
Merged density & Low (318)  & 0.781 & 0.750 & 0.781 & 0.783 \\
Merged density & Medium (265) & 0.792 & 0.759 & 0.796 & 0.786 \\
Merged density & High (10) & 0.735 & 0.725 & 0.725 & 0.725 \\
\midrule
Duplicate-value & None (668) & 0.823 & 0.789 & 0.822 & 0.820 \\
Duplicate-value & Low (246)  & 0.773 & 0.741 & 0.778 & 0.775 \\
Duplicate-value & Medium (98) & 0.841 & 0.792 & 0.846 & 0.837 \\
Duplicate-value & High (11) & 0.829 & 0.782 & 0.836 & 0.833 \\
\bottomrule
\end{tabular}
\caption{Structure-conditioned nEM on STB L1--L5 by merged-cell density and duplicate-value ambiguity. Table counts are shown in parentheses.}
\label{tab:appendix_structure_conditioned}
\end{table}

\subsection{Failure Modes and Question Types}
\label{app:failure_qtype}

Refusal and Invalid rates are low for most formats. LaTeX nevertheless shows substantially higher Invalid and logged-error rates than the other text formats (Table~\ref{tab:appendix_failure_modes}).

\begin{table}[H]
\centering
\small
\begin{tabular}{lrrrr}
\toprule
\textbf{Format} & \textbf{nEM} & \textbf{Refusal} & \textbf{Invalid} & \textbf{Logged error} \\
\midrule
CSV   & 0.808 & 0.22 & 0.09 & 0.11 \\
HTML  & 0.812 & 0.30 & 0.11 & 0.15 \\
LaTeX & 0.777 & 0.38 & 2.67 & 3.55 \\
LTEM  & 0.813 & 0.15 & 0.14 & 0.18 \\
MD    & 0.810 & 0.46 & 0.15 & 0.20 \\
MMD   & 0.810 & 0.34 & 0.15 & 0.19 \\
\bottomrule
\end{tabular}
\caption{Failure diagnostics on STB. Refusal and Invalid are calculated among nEM-incorrect outputs; Logged error is recorded independently and may overlap. Values other than nEM are percentages.}
\label{tab:appendix_failure_modes}
\end{table}

Table~\ref{tab:appendix_question_types} reports descriptive nEM by question type. Aggregation is the most difficult type across formats; we do not infer a format-specific question-type advantage from this breakdown.

\begin{table}[H]
\centering
\small
\begin{tabular}{lrrrr}
\toprule
\textbf{Format} & \textbf{Aggregation} & \textbf{Comparison} & \textbf{Counting} & \textbf{Lookup} \\
\midrule
CSV   & 0.529 & 0.949 & 0.919 & 0.834 \\
HTML  & 0.531 & 0.954 & 0.925 & 0.838 \\
LaTeX & 0.513 & 0.915 & 0.891 & 0.788 \\
LTEM  & 0.530 & 0.954 & 0.925 & 0.843 \\
MD    & 0.529 & 0.955 & 0.923 & 0.831 \\
MMD   & 0.532 & 0.955 & 0.924 & 0.830 \\
\bottomrule
\end{tabular}
\caption{Question-type nEM breakdown on STB.}
\label{tab:appendix_question_types}
\end{table}

\subsection{Fine-tuning Pilot}
\label{app:fine_tuning_pilot}

To assess whether small instruction-tuned models can adapt to LTEM with limited supervision, we conducted a LoRA pilot using LTEM-formatted STB QA examples. Tables were partitioned across the training, validation, and test sets, yielding 3,472, 424, and 460 examples, respectively. We used LoRA rank 16, alpha 32, dropout 0.05, two epochs, a learning rate of $2\times10^{-4}$, and a maximum sequence length of 4096.

Table~\ref{tab:appendix_lora_models} shows substantial gains for all three backbones on held-out STB tables. These results indicate that LTEM-formatted table QA is amenable to adaptation with limited in-domain supervision. Because format exposure and task supervision are introduced jointly, however, the pilot does not isolate their individual contributions. The evaluation is limited to 4B-scale backbones, the STB distribution, and a single set of LoRA hyperparameters.

\begin{table}[H]
\centering
\small
\begin{tabular}{lrrr}
\toprule
\textbf{Backbone} & \textbf{Base nEM} & \textbf{LoRA nEM} & \textbf{$\Delta$ nEM} \\
\midrule
Qwen3-4B-Instruct & 0.489 & 0.763 & +0.274 \\
Phi-4-mini        & 0.539 & 0.737 & +0.198 \\
Gemma-3-4B-IT     & 0.361 & 0.707 & +0.346 \\
\bottomrule
\end{tabular}
\caption{LTEM LoRA pilot results on held-out STB tables using normalized Exact Match (nEM) as defined in the main evaluation.}
\label{tab:appendix_lora_models}
\end{table}

For Qwen3-4B-Instruct, the largest gains occur on comparison and aggregation questions (Table~\ref{tab:appendix_qwen_lora_breakdown}), while lookup and counting improve more modestly.

\begin{table}[H]
\centering
\small
\begin{tabular}{lrrr}
\toprule
\textbf{Slice} & \textbf{Base nEM} & \textbf{LoRA nEM} & \textbf{$\Delta$} \\
\midrule
LOOKUP      & 0.817 & 0.983 & +0.165 \\
COUNTING    & 0.609 & 0.696 & +0.087 \\
AGGREGATION & 0.148 & 0.478 & +0.330 \\
COMPARISON  & 0.383 & 0.896 & +0.513 \\
\midrule
L1 & 0.500 & 0.773 & +0.273 \\
L2 & 0.522 & 0.750 & +0.228 \\
L3 & 0.457 & 0.800 & +0.343 \\
L4 & 0.529 & 0.765 & +0.235 \\
L5 & 0.375 & 0.725 & +0.350 \\
L6 & 0.531 & 0.688 & +0.156 \\
\bottomrule
\end{tabular}
\caption{Qwen3-4B LTEM LoRA breakdown by question type and complexity level.}
\label{tab:appendix_qwen_lora_breakdown}
\end{table}

\section{McNemar's Test for LTEM vs.\ LaTeX on STB}
\label{app:mcnemar}

To assess paired differences between LTEM and LaTeX on STB, we applied McNemar's test to matched predictions within each of three runs and combined the resulting p-values using Fisher's method.

\begin{table}[H]
\centering
\small
\begin{tabular}{lrrrl}
\toprule
\textbf{Model} & \textbf{n$_{01}$} & \textbf{n$_{10}$} & \textbf{Fisher \textit{p}} & \textbf{Winner} \\
\midrule
DeepSeek-R1    & 196 & 343 & 5.4$\times$10$^{-9}$*** & LTEM \\
Llama-4-Maverick & 343 & 441 & 0.0056** & LTEM \\
Gemini-2.5-Pro & 253 & 275 & 0.410 (ns) & --- \\
GPT-5          & 246 & 353 & 2.1$\times$10$^{-4}$*** & LTEM \\
Qwen3-235B     & 398 & 2{,}377 & $<$10$^{-100}$*** & LTEM \\
\bottomrule
\end{tabular}
\caption{McNemar test comparing LTEM vs.\ LaTeX on STB. n$_{01}$ denotes LTEM-wrong \& LaTeX-correct, and n$_{10}$ denotes LTEM-correct \& LaTeX-wrong. The discordant counts are summed across three runs, and the p-values are Fisher-combined. Significance: *** \textit{p}$<$0.001, ** \textit{p}$<$0.01, * \textit{p}$<$0.05.}
\label{tab:mcnemar}
\end{table}

\section{Translation Disclosure}
\label{sec:appendix_translation}
 
We used AI tools as auxiliary aids for some translation. All translations were reviewed and revised by the authors, and the final manuscript reflects our verification of factual accuracy and phrasing.

\end{document}